# GM-Livox: An Integrated Framework for Large-Scale Map Construction with Multiple Non-repetitive Scanning LiDARs

Yusheng Wang, *Graduate Student Member, IEEE*, Yidong Lou, Weiwei Song,
Huan Yu and Zhiyong Tu

*Abstract*—With the ability of providing direct and accurate enough range measurements, light detection and ranging (LiDAR) is playing an essential role in localization and detection for autonomous vehicles. Since single LiDAR suffers from hardware failure and performance degradation intermittently, we present a multi-LiDAR integration scheme in this article. Our framework tightly couples multiple non-repetitive scanning LiDARs with inertial, encoder, and global navigation satellite system (GNSS) into pose estimation and simultaneous global map generation. Primarily, we formulate a precise synchronization strategy to integrate isolated sensors, and the extracted feature points from separate LiDARs are merged into a single sweep. The fused scans are introduced to compute the scan-matching correspondences, which can be further refined by additional real-time kinematic (RTK) measurements. Based thereupon, we construct a factor graph along with the inertial preintegration result, estimated ground constraints, and RTK data. For the purpose of maintaining a restricted number of poses for estimation, we deploy a keyframe based sliding-window optimization strategy in our system. The real-time performance is guaranteed with multi-threaded computation, and extensive experiments are conducted in challenging scenarios. Experimental results show that the utilization of multiple LiDARs boosts the system performance in both robustness and accuracy.

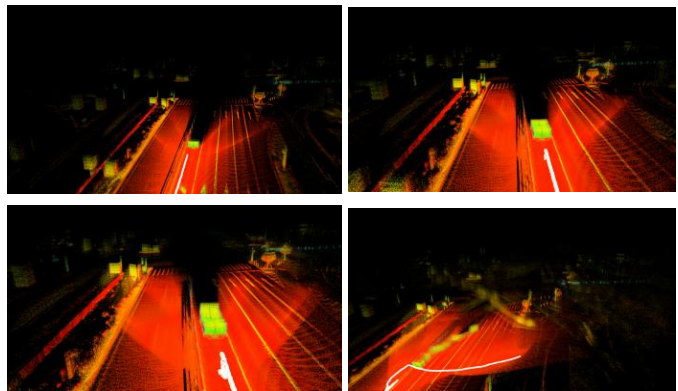

Fig. 1. The mapping failure procedure of a tightly-coupled single LiDAR approach, where the pose estimation is misled by a bus ahead.

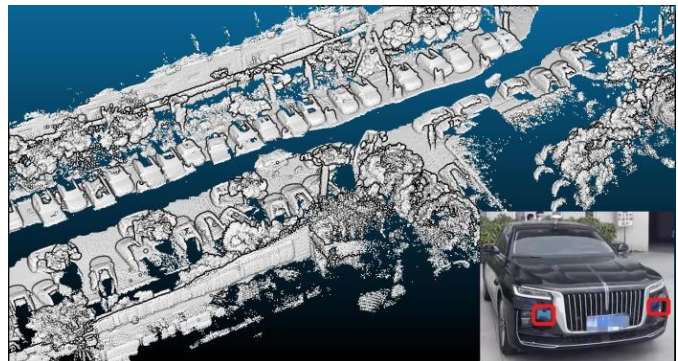

Fig. 2. Visual illustration of the GM-Livox mapping at an outdoor garage, the two red rectangles in the right bottom inset indicates the installed LiDARs on each side. In view of the low mounting height and restricted view angles, all the single Livox LiDAR based approaches fail at such districts.

## I. INTRODUCTION

ROBUST global localization is a fundamental property of building high definition (HD) maps for autonomous driving. Current vehicular positioning schemes are monopolized by global navigation satellite systems (GNSS), for instance, the global positioning system (GPS) and Beidou. Previous studies have reported that the accuracy of single-point positioning (SPP) is about 10 m considering clock errors, satellite orbit, ionospheric and tropospheric delays. These errors can be calibrated out utilizing external correction sources, with precise point positioning (PPP) and real-time kinematic (RTK) two approaches in this category. However, these approaches are limited by GNSS signal quality, and require external sensors to compensate for signal blockage and multipath problems [1]. As a direct measurement of 3D environment and invulnerable to illumination variations, light detection and ranging (LiDAR) is a promising sensor towards this demanding task.

Manuscript submitted July 8, 2021. This work was supported by the Joint Foundation for Ministry of Education of China under Grant 6141A0211907 (*Corresponding author: Yidong Lou*).

Yusheng Wang, Yidong Lou, Zhiyong Tu, Weiwei Song are with the GNSS Research Center, Wuhan University, 129 Luoyu Road, Wuhan 430079, China (email: yushengwhu@whu.edu.cn; ydlouw@whu.edu.cn; sww@whu.edu.cn 2012301650022@whu.edu.cn).

Huan Yu is with the School of Geodesy and Geomatics, Wuhan University, 129 Luoyu Road, Wuhan 430079, China and with INDRIVING Co.,Ltd, Central China Intellectual Valley, Wuhan 430076, China (email: yuhuan@indrv.cn).

The focus of recent LiDAR technology has been on massive deployment in robotic applications, with the solid state LiDAR (SSL) having the greatest concern and potential. Considering the limited reliability and the quantity production capability of SSL at present time, many manufacturers shift to hybrid SSL, and the Livox LiDAR is attracting considerable interest with desirable price and accuracy. Compared with spinning LiDARs, Livox LiDARs all have limited view angles, which will lead to fewer features in a sweep, making the subsequent feature tracking prone to degenerate. Although this problem can be partially solved by tightly-coupled algorithms, single LiDAR solution is still not robust to dynamic objects. From our experience, the LiDAR odometry can be easily misguided by nearby buses or trucks, especially at slow motions (Fig. 1).

To address the aforementioned problems, many vehicles choose to equip additional LiDARs for increased detection range and awareness, such as 2 Livox Hap LiDARs on Xpeng P5 and 4 corner LiDARs on Honda Legend. And the foreground of multiple LiDAR integration has been extensively explained in our previous work [2].

In this paper, we extend our previous work by tightly coupling all the sensor measurement into pose estimation and map construction, namely GM-Livox, to ensure large scale mapping. A mapping example is shown in Fig. 2, where we can tell our proposed algorithm is robust for challenging environments. Compared with existing work, the novelty and main contributions of our work can be concluded as follows:

1) A RTK assisted scan matching method for high speed and fast rotation scenarios.
2) A tightly coupled multiple LiDAR, IMU, encoder, and RTK measurements pose estimation framework, which runs in real time and delivers promising odometry accuracy in large scale and challenging environments.
3) A multisensory ground-constrained mapping method that generates less blurry global map with the majority of dynamic objects been removed.

The rest of the article is organized as follows. Related work is reviewed in Section II. Section III describes the system setup and preparatory works. Section IV formulates the problem and provides general concept. Section V presents the detailed graph optimization process applied in our system, followed by experimental results in Section VI. Finally, Section VII draws conclusions and demonstrates future research directions.

## II. Related Work

The LiDAR odometry is calculated from the scan-to-scan displacement, and can be broadly classified into three different categories: point-based, feature-based and mathematical-property-based scan matching. Since the point-based methods making full use of points from raw LiDAR scans, they can be treated as dense approaches. On the other hand, the feature-based approaches [3], [4] are regarded as sparse methods, as they only employ a select number of points for correspondence tracking. Furthermore, the mathematical-property-based means take advantage of statistical models, and transform the discrete representations from a single scan into a continuous distribution.

Many LiDAR-only simultaneously localization and mapping (SLAM) are variations of point-based iterative closest point (ICP) registration. Robustness is further improved through integration with other measurements, such as IMU [3]–[5], encoder [6] and wireless communication [7]. And this raises the awareness of integration schemes, which can be categorized into the loosely coupled and the tightly coupled ones. There are numerous works on loosely coupled odometry in the literature, where the pose estimations from individual measurements are fused separately. LOAM [3] takes the translation and orientation calculated by the IMU as initial guess for registration, and discards the IMU data for further optimizations. And a loop closure optimization can be added to the back end for correction of the long-term accumulation drift. Since the loosely-coupled systems take the odometry part as a black box and decouple the individual measurements, they are likely to suffer information loss and become less accurate.

In contrast, tightly-coupled methods directly fuse isolated sensor measurements through joint optimization [8], and can be classified into optimization-based and filter-based approaches. LIO-mapping [9] jointly optimizes measurements from lidar and IMU, and the long-during drift can be corrected with detected loops. Based on the incremental smoothing and mapping structure iSAM2, Lio-sam [10] views IMU preintegration, LiDAR odometry and global optimization as different types of factors, and relies heavily on LiDAR odometry to further constrain pre-integrated IMU states in the factor graph. This scheme can result in loss of constraint information posed by landmarks, besides, high-frequency IMU information and precise calibration are also needed for LiDAR de-skewing. A robocentric lidar-inertial estimator LINS is proposed in [11], which recursively corrects the estimated state using an iterated error-state Kalman filter (ESKF). Although achieving superior accuracy with high computation efficiency, this method also suffers from drift during long-term navigation.

According to the data association scheme, multi LiDAR integration can be classified into decentralized and centralized approaches. A decentralized framework of multiple Livox LiDARs is proposed in [12], based on Extended Kalman filter (EKF), this method treats each LiDAR input as independent modules for pose estimation. Although approval accuracy can be reached, this system is only simulated on a high-performance computer, the communication delay and message loss in the real case are not taken into consideration. A centralized multi LiDAR method is presented in [13], this approach runs onboard with pose estimation, online extrinsic refinement, convergence identification, and mapping in the meantime. However, as a LiDAR-only solution, this approach is inevitable to long-duration navigation drifts.

Eight LiDARs are integrated into estimation in our previous work [2], with improved feature selection method and loop detector, it can reach promising result with only IMU measurements. Multi LiDAR integration provides the convenience of manipulating each LiDAR feature points input, which can alleviate the degeneration influence in certain areas. In this view, our method outperforms classical spinning LiDAR based approaches in challenging conditions.

## III. Implementation of Hardware System

We conduct experiments using the detailed hardware setup as shown in Fig. 3. Six Livox Horizon[1] LiDARs are included with a Livox Hub[2] in charge of connection. The IMU and GNSS measurements come from a Femotomes MiniII-D-INS [3] navigation unit, with RTK corrections send from Qianxun SI[4]. The encoder used is a SICK DFS60E-S1CC02000[5].

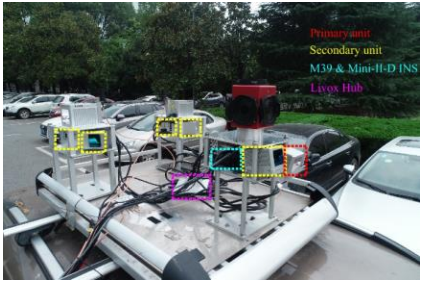

Fig. 3. Experimental setup.

Instead of using the built-in Linux ptp software synchronization method, we apply the hard synchronization method described in Fig. 4(a). All the six LiDARs are synchronized with a u-blox EVK-M8T [6] GNSS timing evaluation kit using GNSS pulse per second (PPS), and the timestamp is replaced with the GPRMC format.

So far, the sensors are precisely configured, and Fig. 4(b) presents an illustration of the synchronization process in the algorithm. As shown in Fig. 3, we select the down view LiDAR as the primary unit, and others are regarded as secondary LiDARs. The scan period of the primary LiDAR is approximately 0.1 s, thus for any $k$, the time-span between $t_k$ and $t_{k+1}$ is also 0.1 s. However, the secondary LiDARs are less likely to have the identical timestamps with the primary unit due to unpredictable time delays and information loss. As they are sent to different threads for feature extraction, we then merge all the feature points whose start time fall in $[t_k, t_{k+1}]$ to obtain fused scan $\mathcal{F}_k$. The combined points inherit the timestamp of the primary LiDAR and stretch over the time span $[t_k, t'_k]$. The IMU measurements in the sub interval $[t_{k-1}, t'_k]$ are used for state propagation, where the samples in $[t_{k-1}, t_k]$ are utilized for preintegration and the other subsets are introduced for motion compensation.

Furthermore, as our system receives raw information from numerous sensors, the built-in transmission control protocol (TCP) in robot operation system (ROS) [14] is susceptible to message loss due to filled buffers. Therefore, we choose the lightweight communications and marshalling (LCM) library for data exchanging, which minimizes latency and maintains high bandwidth as well.

Although the lever-arms of encoder and the two GNSS antennas with respect to the integrated navigation unit can be determined by a total station, the extrinsic of multiple LiDARs as well as the displacements between LiDARs and the navigation unit are hard to configure considering the sophisticated geometry structures.

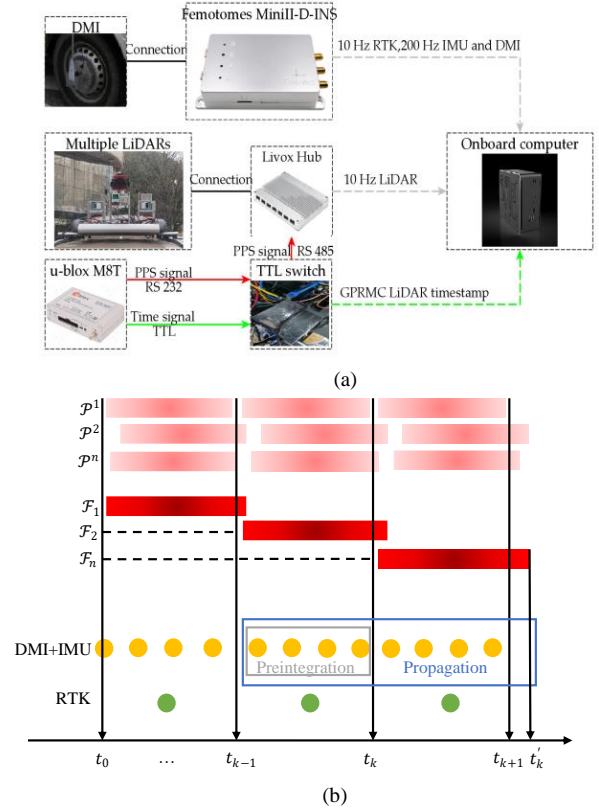

Fig. 4. Illustration of the synchronization process within multiple sensors, (a) is the hardware synchronization framework and (b) is the software scheme.

We hereby adopt the target-free self-calibration method [15] for multiple LiDARs, where each LiDAR independently generate the local map along the path then the generalized-ICP [16] is used for map alignment. As illustrated in our previous work, this method can achieve millimeter-level accuracy after several trials. Then the point clouds of six Horizon LiDARs can be merged into a big scan, roughly centered at the geometric center of the roof platform.

With the assumption of all the sensors been rigidly installed, we exploit the motion-based solution to solve the LiDAR-GNSS mounting parameters. This method utilizes the independent motion estimation result of different sensors to find the transformation relations between them. We hereby choose an open area with rich features as the calibration site, and employ the multi non-repetitive scanning LiDAR SLAM [2] to compute the LiDAR motion, this information is then aligned with RTK measurements. Moreover, large range of motions is satisfied through a ∞ driven pattern, ensuring an accurate calibration result.

---

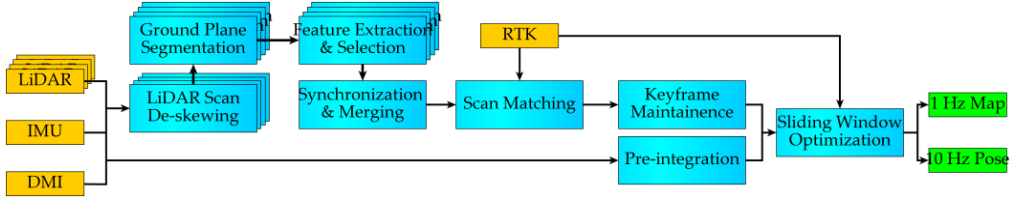

Fig. 5 . System overview

## IV. SYSTEM FRAMEWORK

We first define frames and notations that are used throughout this paper. The notation $(\hat{\cdot})$ denotes the noisy measurement. We use both rotation matrices $\boldsymbol{R} \in SO(3)$ and quaternions $\boldsymbol{q}$ to represent rotation. The vehicle body frame with the IMU center as the origin is denoted as $(\cdot)^B$ frame, the LiDAR frame is defined as $(\cdot)^L$, the encoder frame is $(\cdot)^O$, the reference world frame is $(\cdot)^W$. In addition, we define $(\cdot)^B_W$ as the transformation from $W$ to $B$ and $(\cdot)^{OB}$ as the encoder result expressed in the $B$ frame. The robot state $\boldsymbol{x}$ can then be written as:

$$\boldsymbol{x} = [\boldsymbol{R}^T, \boldsymbol{p}^T, \boldsymbol{v}^T, \boldsymbol{b}_a^T, \boldsymbol{b}_g^T, s_o]^T \quad (1)$$

where $\boldsymbol{p} \in \mathbb{R}^3$, $\boldsymbol{v} \in \mathbb{R}^3$ are respectively the position and linear velocity vector. $\boldsymbol{b}_a$ and $\boldsymbol{b}_g$ are the usual IMU gyroscope and accelerometer biases, $s_o$ is the encoder scale factor.

The proposed GM-Livox scheme is shown in Fig. 5 . The system receives sensor data from multiple Livox LiDARs, an IMU, encoder and optionally RTK measurements. The problem of crosstalk is an important consideration when connecting multiple sensors, thus we first remove the noisy points and synchronize the raw point clouds with the IMU. For each input LiDAR, a sequence of point clouds can be obtained, and they are de-skewed using IMU data. After the ground segmentation process, the edge features are extracted from the remaining points, and all the feature points are synchronized and merged into a fused point cloud. We seek to estimate the vehicle ego-motion and keep a globally consistent map simultaneously, which can be formulated into a maximum a posterior (MAP) problem. We use a factor graph to model this problem, with encoder assisted IMU preintegration factors, LiDAR factors and RTK factors introduced for graph construction.

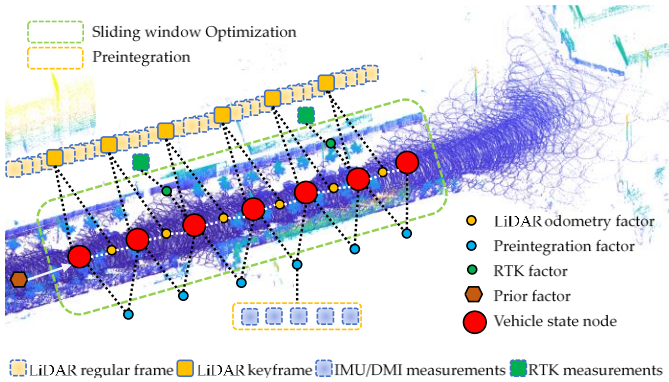

Fig. 6. Proposed fusion scheme.

## V. TIGHTLY COUPLED FUSION SCHEME VIA KEYFRAME BASED SLIDING WINDOW OPTIMIZATION

As an intuitive way of formulating SLAM problem, graph-based optimization uses nodes to represent the poses of the vehicle at different points in time. A new node is added to the graph when the pose displacements exceed a certain threshold, then the factor graph is optimized upon the node insertion. For the sake of increasing computation efficiency, we employ the sliding window optimization to keep a relative steady number of nodes in the local graph as shown in Fig. 6.

For a sliding window containing $k$ keyframes, the optimal keyframe states $\boldsymbol{X} = [\boldsymbol{x}_1^T, \boldsymbol{x}_2^T, \ldots, \boldsymbol{x}_k^T]^T$ can be acquired through minimizing:

$$\min_{\boldsymbol{X}} \{ \|\boldsymbol{r}_p\|^2 + \sum_{i=1}^{k} \|\boldsymbol{r}_{\mathcal{J}_i}\|^2 + \sum_{i=1}^{k} \boldsymbol{r}_{\mathcal{L}_i} + \sum_{i=1}^{k} \|\boldsymbol{r}_{\mathcal{G}_i}\|^2 + \sum_{i=1}^{k} \|\boldsymbol{r}_{\mathcal{P}_i}\|^2 \} \quad (2)$$

where $\boldsymbol{r}_p$ is the prior factor marginalized by Schur-complement [17], $\boldsymbol{r}_{\mathcal{J}_i}$ is the residual of IMU/encoder preintegration. $\boldsymbol{r}_{\mathcal{L}_i}$ and $\boldsymbol{r}_{\mathcal{G}_i}$ defines the residual of relative LiDAR constraints and the ground constraints, respectively. Finally, the residual of global positioning system is $\boldsymbol{r}_{\mathcal{P}_i}$.

### A. Preintegration Factor

The pre-integrated IMU/odometer residuals from $k$ to $k+1$ is denoted as $\hat{\boldsymbol{z}}_{k+1}^k = [\hat{\boldsymbol{\alpha}}_{B_{k+1}}^{B_k}, \hat{\boldsymbol{\beta}}_{B_{k+1}}^{B_k}, \hat{\boldsymbol{\gamma}}_{B_{k+1}}^{B_k}, \hat{\boldsymbol{\alpha}}_{OB_{k+1}}^{OB_k}]^T$, representing variations in position, velocity, attitude and odometer increment. In the discrete time case, the preintegration terms can be computed recursively using:

$$\hat{\boldsymbol{\alpha}}_{B_{i+1}}^{B_k} = \hat{\boldsymbol{\alpha}}_{B_i}^{B_k} + \hat{\boldsymbol{\beta}}_{B_i}^{B_k}\delta t + \frac{1}{2} \boldsymbol{R}(\hat{\boldsymbol{\gamma}}_{B_i}^{B_k})(\hat{\boldsymbol{a}}_i - \boldsymbol{b}_{a_i})\delta t^2$$
$$\hat{\boldsymbol{\beta}}_{B_{i+1}}^{B_k} = \hat{\boldsymbol{\beta}}_{B_i}^{B_k} + \boldsymbol{R}(\hat{\boldsymbol{\gamma}}_{B_i}^{B_k})(\hat{\boldsymbol{a}}_i - \boldsymbol{b}_{a_i})\delta t$$
$$\hat{\boldsymbol{\gamma}}_{B_{i+1}}^{B_k} = \hat{\boldsymbol{\gamma}}_{B_i}^{B_k} \otimes \begin{bmatrix} 1 \\ \frac{1}{2}(\hat{\boldsymbol{\omega}}_i - \boldsymbol{b}_{\omega_i})\delta t \end{bmatrix} \quad (3)$$
$$\hat{\boldsymbol{\alpha}}_{OB_{i+1}}^{OB_k} = \hat{\boldsymbol{\alpha}}_{OB_i}^{OB_k} + \boldsymbol{R}(\hat{\boldsymbol{\gamma}}_{B_i}^{B_k})s_{OB_{i+1}}^{OB_i}$$

where $\boldsymbol{R}(\hat{\boldsymbol{\gamma}}_{B_i}^{B_k})$ is the rotation matrix representation of $\hat{\boldsymbol{\gamma}}_{B_i}^{B_k}$, $\hat{\boldsymbol{a}}_i$ and $\hat{\boldsymbol{\omega}}_i$ are the raw gyroscope and accelerometer measurements denoted in [18], $\boldsymbol{s}_{OB_{i+1}}^{OB_i}$ is the odometer increment measurement. Thus the residuals of the preintegration can be defined as:

$$\boldsymbol{r}_{\mathcal{J}_k} = \begin{bmatrix} \boldsymbol{R}_W^{B_k}(\boldsymbol{p}_{B_{k+1}}^W - \boldsymbol{p}_{B_k}^W + \frac{1}{2}\boldsymbol{g}^W \Delta t_k^2 - \boldsymbol{v}_{B_k}^W \Delta t_k) - \hat{\boldsymbol{\alpha}}_{B_{k+1}}^{B_k} \\ \boldsymbol{R}_W^{B_k}(\boldsymbol{v}_{B_{k+1}}^W + \boldsymbol{g}^W \Delta t_k - \boldsymbol{v}_{B_k}^W) - \hat{\boldsymbol{\beta}}_{B_{k+1}}^{B_k} \\ 2\left[(\boldsymbol{q}_{B_k}^W)^{-1} \otimes (\boldsymbol{q}_{B_{k+1}}^W) \otimes (\hat{\boldsymbol{\gamma}}_{B_{k+1}}^{B_k})^{-1}\right]_{2:4} \\ \boldsymbol{b}_{a_{k+1}} - \boldsymbol{b}_{a_k} \\ \boldsymbol{b}_{g_{k+1}} - \boldsymbol{b}_{g_k} \\ \boldsymbol{R}_W^{B_k}(\boldsymbol{p}_{B_{k+1}}^W - \boldsymbol{p}_{B_k}^W) - \hat{\boldsymbol{\alpha}}_{OB_{k+1}}^{OB_k} \\ s_{O_{k+1}} - s_{O_k} \end{bmatrix} \quad (4)$$

where $\boldsymbol{g}^W = [0, 0, g]^T$ is the gravity factor in the world frame and $\Delta t_k$ is the time sweep between $k$ and $k+1$. We use $[\cdot]_{2:4}$ to take out the last three elements of a quaternion vector.

*B. LiDAR Odometry Factor*

Considering the large variance in the direction of beams, we first remove the point clouds within the range of 3 m w.r.t the separate LiDARs. Then we use the IMU/encoder state propagation to compensate for the point cloud distortion with quaternion-based spherical linear interpolation operation.

Inspired by [4], we perform a ground point segmentation for effective and accurate feature correspondences estimation. We use the LiDAR mounting height and laser beam angel to identify the points hitting on ground, and a random sample consensus (RANSAC) [19] algorithm is applied for reliable plane fitting. The planar features are only extracted from the ground points and the edge features are extracted from the remaining points all through smoothness evaluation similar with [3]. Since Livox LiDAR is sensitive to reflectivity changes (e.g., road curbs and lane markings are clearly distinguishable), we also employ the intensity as an extra determinant for feature extraction. In addition, we exploit a feature selection algorithm in [2] to improve efficiency and avoid degeneration. Then we take the fused feature points $\mathcal{F}_k$ to perform scan registration with the edge and planar patch correspondence computed using the following equation:

$$\boldsymbol{d}_{\mathcal{E}_k} = \frac{\left\|(P_k^W - \epsilon_1^W) \times (P_k^W - \epsilon_2^W)\right\|}{\left\|\epsilon_1^W - \epsilon_2^W\right\|} \quad (5)$$

$$\boldsymbol{d}_{\mathcal{S}_k} = \frac{\left|(P_k^W - \eta_1^W)^T \left((\eta_1^W - \eta_2^W) \times (\eta_1^W - \eta_3^W)\right)\right|}{\left|(\eta_1^W - \eta_2^W) \times (\eta_1^W - \eta_3^W)\right|} \quad (6)$$

here $P_k^W = R_k P_k^L + \boldsymbol{p}_k$ represents the scan point in the global frame. $(\epsilon_1^W, \epsilon_2^W)$ and $(\eta_1^W, \eta_2^W, \eta_3^W)$ are the edge and planar points, respectively. Dynamic objects can be partially filtered out by removing one quarter largest residuals from equation (5) and equation (6). With the number of edge and planar correspondences $n_\epsilon$ and $n_\eta$ in a sweep, the residual of LiDAR constraint can be expressed as:

$$r_{\mathcal{L}_k} = \sum_{k=1}^{n_\epsilon} (\boldsymbol{d}_{\mathcal{E}_k})^2 + \sum_{k=1}^{n_\eta} (\boldsymbol{d}_{\mathcal{S}_k})^2 \quad (7)$$

Additionally, the ground plane $\boldsymbol{m}$ can be parameterized by the united normal direction vector $\boldsymbol{n}_p$ and a distance scalar $d_p$, $\boldsymbol{m} = [\boldsymbol{n}_p^T, d_p]^T$. Then the correspondence of each ground point between two consecutive scan $k$ and $k+1$ can be established by:

$$\mathcal{G}_{k+1} = T_{L_{k+1}}^{L_k} \mathcal{G}_k \quad (8)$$

$$T_{L_{k+1}}^{L_k} = T_B^L (T_{B_k}^W)^{-1} T_{B_{k+1}}^W (T_B^L)^{-1} \quad (9)$$

where $\mathcal{G}_{k+1}$ and $\mathcal{G}_k$ is the same point in frame $L_{k+1}$ and $L_k$, respectively. Based thereon, the ground plane measurement residual can be expressed as:

$$r_{\mathcal{G}_k} = \boldsymbol{m}_{k+1} - T_{L_{k+1}}^{L_k} \boldsymbol{m}_k \quad (10)$$

We notice that wrong matchings often happen at fast turnings or with large velocities. Described in Fig. 7, irregular motions appear in fast turnings and "hesitation" in localization shows in high-speed cases. The above two problems are mainly because of the limited view angles and detection ranges, as insufficient correspondences are prone to degenerate at such areas. Based thereon, we add the RTK/IMU measurements for verification, once the difference between scan registration and RTK/IMU exceeds a certain threshold, the scan registration is further refined with this information (Fig. 8).

Real-time performance is grudgingly obtainable when integrating every LiDAR sweep for computing and adding factors to the graph, thus we embrace the keyframes for maintaining the sparsity of the optimization scheme. As shown in Fig. 6, a regular LiDAR frame is selected as a new keyframe when the divergence of vehicle pose surpass a typical threshold compared with the previous one. The newly inserted keyframe is correlated to a new vehicle state node, and the regular frames within two keyframes are discarded for further optimization. Therefore, it is a vital issue to mitigate the LiDAR odometry drift between two keyframes with restricted frame intervals. We set the position and rotation change threshold for adding keyframes as 10 m and 2°, empirically.

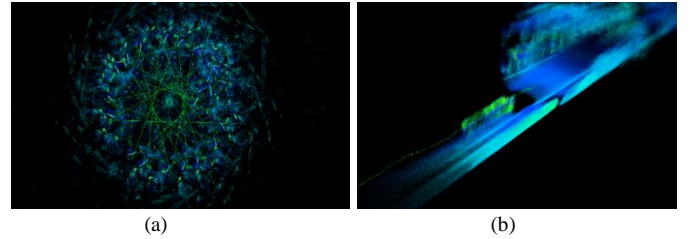

(a)    (b)

Fig. 7. Visual illustration of two registration failure, (a) happens at fast turnings and (b) is because of large velocity.

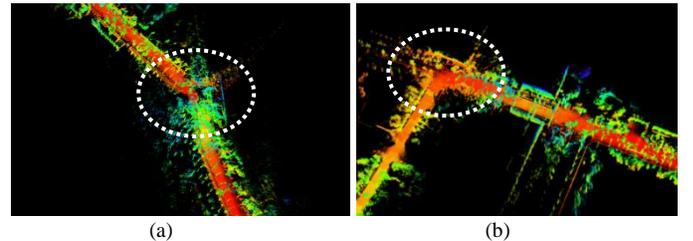

(a)    (b)

Fig. 8. Visual illustration of the RTK aided scan registration. The white dotted ellipse in (a) and (b) denotes the same crossing, RTK measurements is only added as a way of correction in (b).

*C. RTK Factor*

The accumulated drifts of the system can be corrected using RTK measurements. The RTK factor is added when the estimated position covariance is larger than the reported RTK covariance in [10]. However, we find that the reported RTK covariance is not trustworthy sometimes, and may yield blurred or inconsequent mapping result. We hereby model the RTK measurements $\boldsymbol{p}_k^W$ with additive noise, and the global position residual can be defined as:

$$r_{\mathcal{P}_k} = R_W^{B_k} \left(\hat{\boldsymbol{p}}_k^W - \boldsymbol{p}_W^B - \boldsymbol{p}_{B_k}^W + \frac{1}{2} \boldsymbol{g}^W \Delta t_k^2 - \boldsymbol{v}_{B_k}^W \Delta t_k\right) - \hat{\boldsymbol{\alpha}}_{B_{k+1}}^{B_k} \quad (11)$$

where $\hat{\boldsymbol{p}}_k^W$ is the noisy measurement, with $\hat{\boldsymbol{p}}_k^W = \boldsymbol{p}_k^W + \boldsymbol{n}_\mathcal{P}$, here $\boldsymbol{n}_\mathcal{P} \sim \mathcal{N}(\boldsymbol{0}, \sigma_\mathcal{P}^2 \cdot \boldsymbol{I})$ is the Gaussian noise. $\boldsymbol{p}_W^B$ is the transformation from the receiver antenna to the IMU, which can be obtained from installation configuration.

## VI. EXPERIMENTAL RESULTS

All the data is captured and processed by an onboard computer, with Intel i9-10980HK CPU, 64-GB RAM. Besides, all our algorithms are implemented in C++ and executed in Ubuntu Linux using the ROS. The nonlinear optimization and factor graph optimization problem is solved using Ceres Solver and GTSAM [20], respectively. Our proposed system can reach real-time performance for all the captured datasets.

The ground truth is kept by the post-processing result from a MPSTNAV M39 GNSS/INS integrated navigation system [7], which is done by a software [8] similar with Novatel Inertial Explorer, and Table 1 gives the relative positioning performance during GNSS outages.

We conduct extensive evaluations using self-recorded datasets, and four novel Livox Horizon based methods are selected for comparison: a LiDAR-only Livox Mapping [9], a loosely coupled Livox Horizon Loam [10], an optimization-based Lili-om [21] and a filter-based Fast-lio [22]. Considering the selected methods only employ LiDAR and IMU, we also add the multiple LiDAR and IMU only experiments for ablation study, denoted as M-Livox.

Table 1. Positioning performance of the M39 during GNSS outages.

| Sensor | Method | Outage Duration | Position Horizontal | Position Vertical | Attitude Heading |
|---|---|---|---|---|---|
| M39 | Post Processing | 10 s | 0.12 m | 0.09 m | 0.02° |
|  |  | 30 s | 0.24 m | 0.14 m | 0.03° |
|  |  | 60 s | 0.31 m | 0.16 m | 0.03° |

### A. Long Bridge Dataset

The very first experiment aims to present the superiority of the multi-sensory integration. Presented in Fig. 9, we choose the long bridge dataset for demonstration here. The journey is about 3.7 km long with an average speed of 65 km/h. Since the feature points are mostly extracted from lamps and the fan-design bridge cables, the LiDAR odometry cannot work well at such districts due to feature degeneration, and the four selected algorithms all fail to produce meaningful results. M-Livox is also dead as simply adding LiDAR amounts cannot deal with repetitive structures. Thanks to the wheel encoder and RTK information, GM-Livox finishes the job and maintains a low positioning error as well.

Although the direct georeferencing method applied on mobile mapping system (MMS) sounds great for this open-air dataset, it over relies on the precision of RTK/INS measurement, and a sharp vibration will result in pavement aliasing illustrated in the inset 1 of Fig. 9. Such vibrations are also harmful to system with directly-inserted global positioning measurements [10], which deeply depends on the reported GNSS covariance.

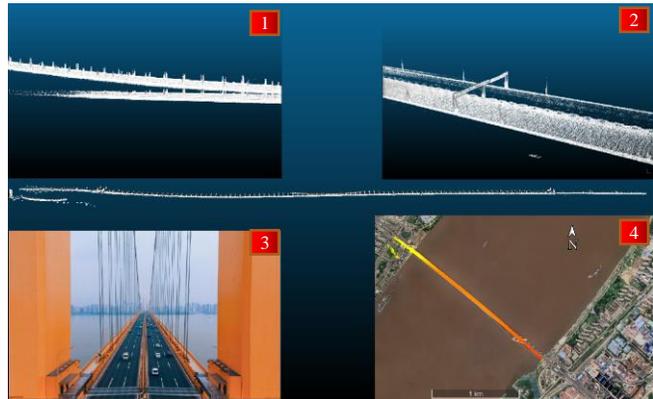

Fig. 9. The long bridge mapping result of GM-Livox. The middle is the vertical view of the GM-Livox mapping result. Inset 1 denotes the result from direct georeferencing, inset 2 is the zoom in of the GM-Livox mapping. Inset 3 is the image of the bridge, and inset 4 is the GM-Livox mapping aligned with Google Earth, the color indicates height variations.

### B. Dong Lake Dataset

This experiment is designed to reveal the advantage of our proposed system against novel single LiDAR approaches. Presented in Fig. 10(a), the overall length of Dong Lake dataset is around 4.3 km with an average speed of 53 km/h, and the RTK status is always at fixed point solution along the path.

Livox Mapping fails at the beginning due to wrong registration caused by dynamic objects. As shown in Fig. 10(b), Livox Horizon Loam also encounters several failures, and the odometry result even moves backwards at the feature-poor areas, where only the road and lamp posts are observable.

Although highly accurate on slow-driven unmanned ground vehicle (UGV), Lili-om encounters a serious heading divergence without additional feature constraints from side-view. This phenomenon is similar with monocular vision based visual-inertial odometry, where the heading errors are unavoidable especially at fast tunings. In addition, the error also accumulates in the roll direction, making a certain slope to the road surface parallel to the vehicle.

With extra information from surround view LiDARs, the observed heading divergence is inconspicuous for M-Livox, but the horizontal displacement is still unsatisfactory. On account of increased sensing coverage and incorporated absolute position measurements, GM-Livox preserves more environmental details and retains a low-level positioning error at the same time. In addition, we plot the absolute error over distance in Fig. 10(c) for detailed reference.

We list some of the mapping advantages of our system in Fig. 11. It can be noticed in Fig. 11(a) that the mapping result of Lili-om is blurred. The traffic sign board is misaligned, and three rows of pipes appeared, whereas there are only two of them in reality. Besides, the lampposts are elongated with some noisy points. We believe these two errors are mainly because of inaccurate pose estimations, and they can be eliminated with increased sensor integrations. In addition, most of the moving vehicles are removed in Fig. 11(b), which is a crucial feature for map constructions.

---

[7] http://www.whmpst.com/cn/imgproduct.php?aid=80
[8] http://www.whmpst.com/en/imgproduct.php?aid=29
[9] https://github.com/Livox-SDK/livox_mapping
[10] https://github.com/Livox-SDK/livox_horizon_loam

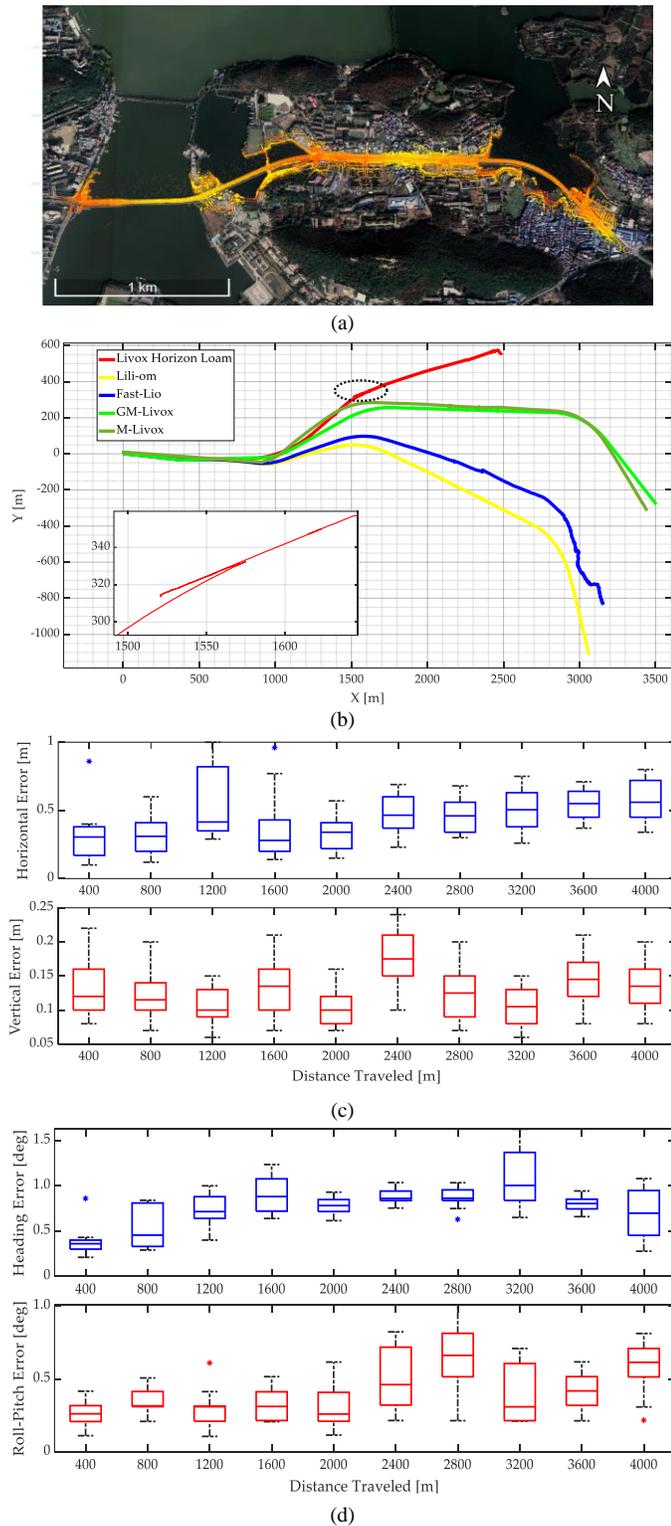

(a)

(b)

(c)

(d)

Fig. 10. Results of various methods using the Dong Lake dataset. (a) is the proposed GM-Livox map aligned with Google Earth, (b) is the visual comparison of trajectories, and the inset denotes the Livox Horizon Loam failure due to degeneration. (c) and (d) are the absolute translational and rotational errors over distance of GM-Livox, respectively.

## C. Bayi Road Dataset

In this experiment, we focus on the system performance when GNSS is temporarily unavailable, and the Bayi road dataset is chosen for illustration here. The journey is 2.3 km in all, with an 850 m tunnel in the middle. As shown in Fig.12(a), our system is robust towards long, featureless tunnel scenario, and the heading error maintains well at the tunnel exit.

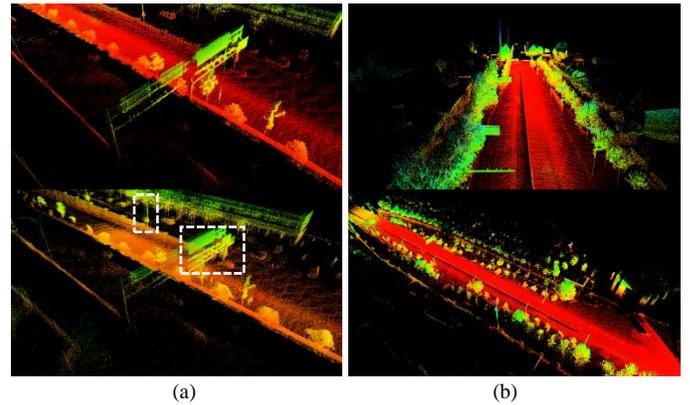

(a)          (b)

Fig. 11. Visual illustration of the advantages of our proposed algorithm. (a) is the comparison against Lili-om, with the above one from our method and the bottom one from Lili-om, the two white dashed rectangles denote some of the evident errors. (b) presents the ability of filtering dynamic objects.

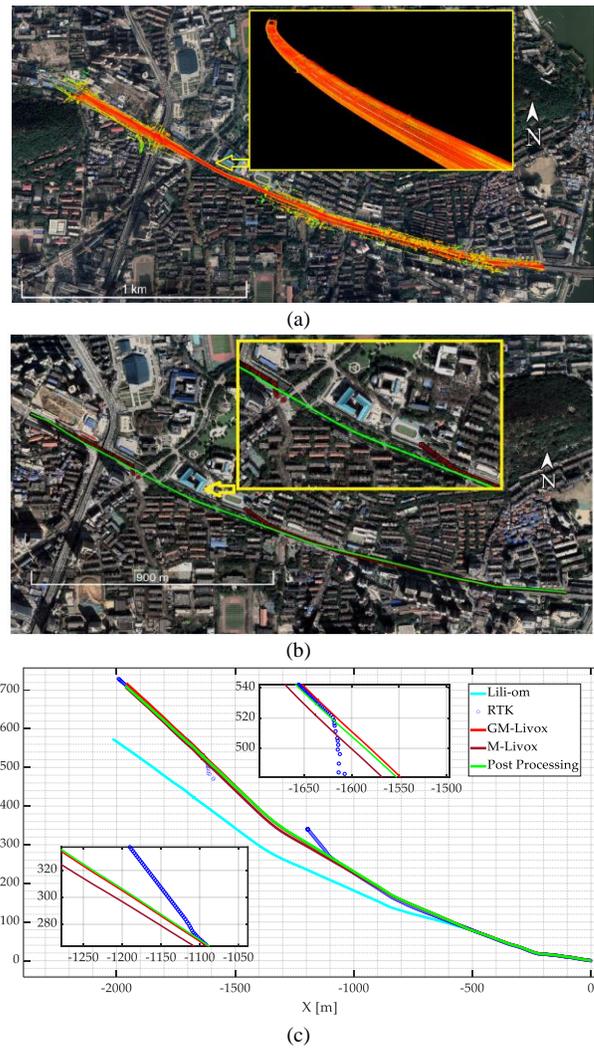

(a)

(b)

(c)

Fig. 12. Results of various methods using the Bayi road dataset. (a) is the proposed GM-Livox map aligned with Google Earth, and the inset in the yellow rectangle indicates the real-time output from GM-Livox. (b) is the proposed GM-Livox trajectory aligned with Google Earth, (c) is the comparison of various methods against ground truth.

Livox Mapping and Livox Horizon Loam "stops" at the tunnel entrance due to wrong registration of repetitive features. Fast-lio moves backwards at the middle of the tunnel, and fails to produce meaningful result. Therefore, we discard them for further assessment.

Depicted in Fig. 12(c), Lili-om encounters a heading deviation owing to the misguidance of a passing-by bus, and the error propagates to the successive pose estimations.

M-Livox and GM-Livox, on the other hand, is immune to the dynamic objects and has a slower drift rate than the RTK trajectory during GNSS outages. We list the root mean square (RMS) error of RTK as well as GM-Livox at tunnel entrance and exit in Fig. 12(c), respectively. Each scenario has a time-span of 30 s, equally divided into two parts, inside and outside the tunnel. It can be inferred from Fig. 12(c) that the IMU and encoder preintegration factors effectually constrain lateral and vertical drifting, maintaining a favorable heading estimation.

Table 2. Root mean square (RMS) error statistics w.r.t ground truth.

|  |  | GM-Livox | RTK |
|---|---|---|---|
| **Into Tunnel** | Horizontal (m) | 0.36 | 3.13 |
|  | Vertical (m) | 0.22 | 3.46 |
|  | Heading (°) | 1.52 | 29.65 |
| **Leave Tunnel** | Horizontal (m) | 0.68 | 2.32 |
|  | Vertical (m) | 0.39 | 2.27 |
|  | Heading (°) | 2.87 | 34.90 |

*D. City-campus Dataset*

In this experiment, we seek to fully explore the system potential in complicated environments, and the city-campus dataset is chosen for demonstration. The vehicle starts in the campus at the left bottom corner in Fig. 13(a), and drives out of the school, finally returns to the start point. The overall length is 5.7 km, with the time consumption of 1657 s.

Depicted in Fig. 13(a), the fixed solution state of RTK reception is intermittent due to the buildings and viaducts overhead. Without the assistance of other sensors, the global position is not accurate at such areas.

Denoted in Fig. 14, mapping is a challenging task for several reasons. Some of the campus paths are surrounded by construction sites with the siding walls set up on each side, providing fewer useful edge features. This problem easily leads to feature degeneracy, and the LiDAR-only and loosely coupled system often fails at such districts. Besides, since the campus locates in the city center, serious traffic jam is encountered with uncountable dynamic objects along the path. The features of moving objects are not stable, as they will induce registration errors. Considering the restricted view angles of a single LiDAR, where mainly the objects on the road are observable, this issue is fatal even for novel tightly coupled algorithms. On account of these problems, the selected four algorithms together with M-Livox all fails to provide meaningful results.

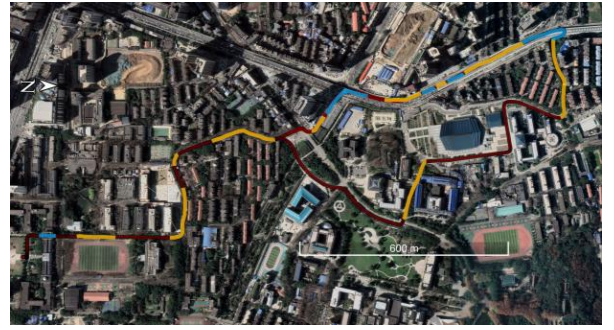

(a)

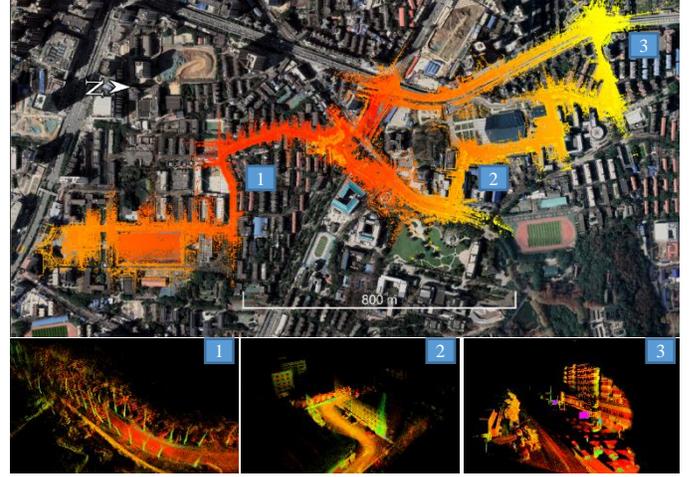

(b)

Fig. 13. The trajectory and mapping result of the city-campus dataset. (a) denotes the vehicle driven path, with red, yellow, and blue indicate fixed solution, float solution, and non-solution of RTK status, respectively. (b) is the mapping result aligned with Google Earth, whereas inset 1 is the campus trail, inset 2 is the city canyon, and inset 3 denotes the traffic jam in the city center.

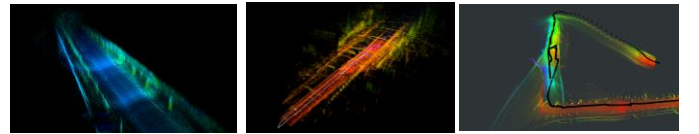

(a)

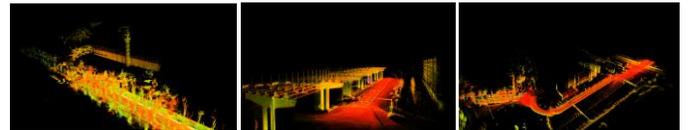

(b)

Fig. 14. Examples of mapping failure of the selected methods. From left to right, (a) denotes the Livox Mapping and Livox Horizon Loam failure due to degeneracy, Lili-om failure due to dynamic objects, fast-lio failure due to fast motions, respectively. (b) is the mapping result of GM-Livox at the corresponding districts.

The proposed GM-Livox system comprehensively exploits multiple LiDARs, IMU, encoder, and RTK measurements in a tightly coupled manner, which accurately estimates the vehicle pose and maintains a global map concurrently. It can be noted from Fig. 15(a) that the estimated trajectory precisely overlaid with the ground truth, with the RMS of the position error 0.41 m. In addition, we plot the position errors in Fig. 15(b) as detailed reference.

We can infer from Fig. 15(b) that the drift is corrected periodically by the inserted RTK data. The vehicle encounters a serious traffic jam under the viaduct during the timespan from 847 s to 1253 s, and the positioning errors grows steadily within this period. Specifically, the RMS of the position error within this period is 0.74 m. After receiving fixed solution RTK data, the accumulated errors are eliminated. Furthermore, thanks to the preintegration results and ground constraints, the Z direction has the most favorable accuracy among the three axes.

Finally, we want to reveal the GM-Livox localization and mapping advantage against mechanical LiDARs, and the 16 beams Velodyne Puck, 64 beams Hesai Pandar64[11], 128 beams Robosense RS-Ruby[12] are chosen for illustration here. We add additional factors to SC-A-LOAM[13], SC-LeGO-LOAM[14] and Lio-sam to make an impartial comparison. We can infer from Fig. 15 that the positioning accuracy has a noticeable promotion with the increased point cloud density, where the system is unable to handle traffic congestion under viaducts with Puck. The multiple Livox LiDAR fusion can reach equivalent accuracy with RS-Ruby as shown in Table 3, and we plot the respective mapping result in Fig. 16. Six Livox LiDAR mapping has an approximal horizontal coverage with that of RS-Ruby, but a remarkable larger vertical coverage, with denser tree crowns.

## VII. CONCLUSION

This article has proposed an integrated GNSS/IMU/encoder pose estimation framework with multiple LiDARs. Experimental results show the validity and robustness of our system in large-scale and challenging environments is encouraging. Although this article is designed for Livox LiDARs, it can be extended to various types of LiDAR combinations, and we have tested on mechanical LiDARs and MEMS LiDARs, for instance.

Recently, many passenger-vehicles are pre-installed with multiple LiDARs to increase sensing coverage and accuracy, and there are also some motor corporations choose Livox LiDARs. Based thereupon, we seek to augment the visual mapping approaches with multiple LiDARs in the future work. One typical application is the autonomous valet parking (AVP), where only the visual semantic information is introduced for mapping and odometry, this solution can only reach 4 m odometry accuracy in the open parking lots [23]. Therefore, this solution is unable to handle high level AVP tasks, where the vehicle should be capable of parking into a specified spot. For comparison, our test with 8 surround view cameras and 3 LiDARs can reach decimeter-level accuracy. Another application is the lightweight HD maps building in urban scenarios. Introducing LiDARs into semantic mapping will lead to favorable improvements.

Table 3. RMS of horizontal localization error statistics for various methods, unit: m.

|  | SC-A-LOAM | SC-LeGO-LOAM | Lio-sam | GM-Livox |
|---|---|---|---|---|
| Puck | / | / | / | / |
| Pandar64 | 4.35 | 5.79 | 2.12 | / |
| RS-Ruby | 2.40 | 1.98 | **0.33** | / |
| Livox | / | / | / | 0.41 |

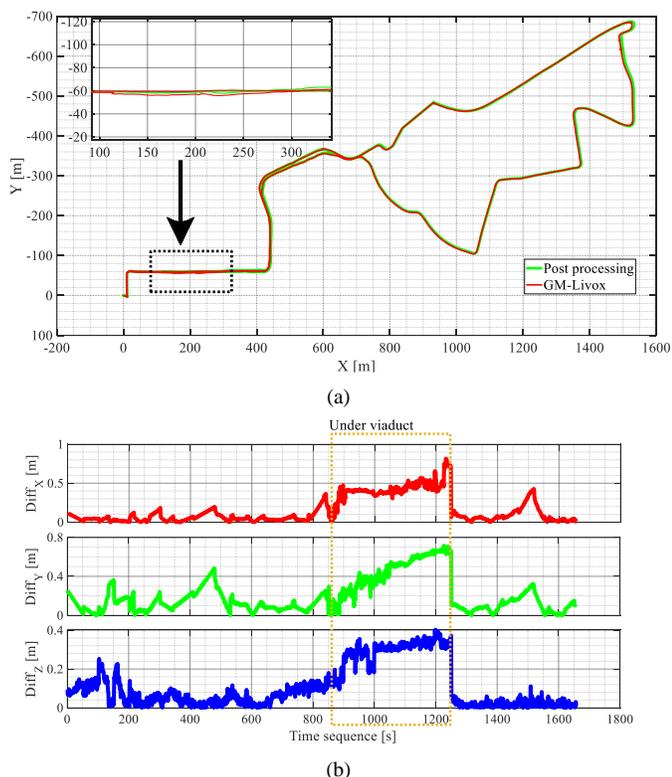

Fig. 15. Evaluation of the GM-Livox odometry result. (a) is the estimated trajectory compared with ground truth, (b) is the detailed position error curve on each axis.

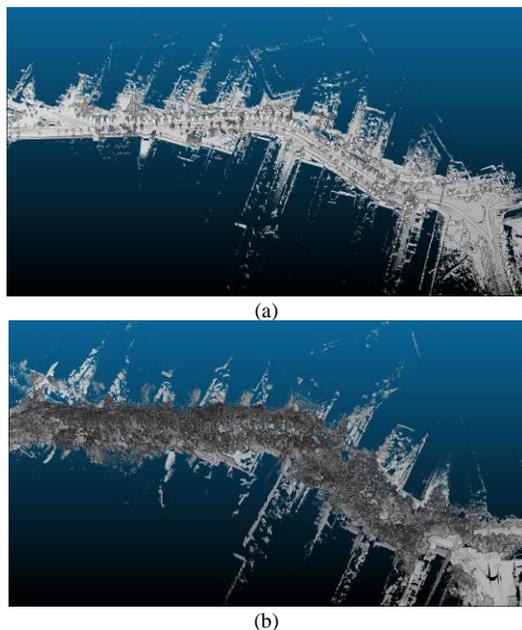

Fig. 16. Different mapping result of inset 1 in Fig. 13(b), with (a) from robosense RS-Ruby and (b) from multiple Livox LiDARs.

---

[11] https://www.hesaitech.com/en/Pandar64
[12] https://www.robosense.ai/en/rslidar/RS-Ruby
[13] https://github.com/gisbi-kim/SC-A-LOAM
[14] https://github.com/irapkaist/SC-LeGO-LOAM